\documentclass[letterpaper, 10 pt, conference]{IEEEtran}  

\IEEEoverridecommandlockouts                              

\usepackage{graphicx}
\usepackage{xcolor}
\usepackage{bm}
\usepackage{amsmath,amssymb,mathrsfs,amsthm,bm}
\usepackage{tikz}
\usepackage{cite}
\usepackage{soul}
\usepackage{subfigure}
\usepackage{tabularx}
\usepackage[normalem]{ulem}
\usepackage[numbers, sectionbib]{natbib}
\usepackage[bookmarks=true]{hyperref}
\usepackage{xurl}
\usepackage{mathtools}
\usepackage{flushend}

\DeclareMathOperator*{\argmax}{arg\,max}

\usepackage{algorithm}
\usepackage{algpseudocode}
\algdef{S}[FOR]{ForEach}[1]{\algorithmicforeach\ #1\ \algorithmicdo}
\algnewcommand\algorithmicinput{\textbf{Input:}}
\algnewcommand\INPUT{\item[\algorithmicinput]}
\algnewcommand\algorithmicoutput{\textbf{Output:}}
\algnewcommand\OUTPUT{\item[\algorithmicoutput]}
\algrenewcommand\algorithmicindent{1.0em}%

\usepackage{eso-pic}
\newcommand\AtPageUpperMyright[1]{\AtPageUpperLeft{%
 \put(\LenToUnit{0.5\paperwidth},\LenToUnit{-1.5cm}){%
     \parbox{0.5\textwidth}{\raggedright\fontsize{11}{11}\selectfont #1}}}}
\newcommand{\conf}[1]{\AddToShipoutPictureBG*{\AtPageUpperMyright{#1}}}

\title{\LARGE \bf
Coordination of Bounded Rational Drones through Informed Prior Policy
}

\author{Durgakant Pushp, Junhong Xu, and Lantao Liu 
\thanks{\newline 
    D. Pushp, J. Xu, and L. Liu are with the Luddy School of Informatics, Computing, and Engineering  at Indiana University, Bloomington, IN 47408, USA. \newline
    E-mail:        {\tt\small \{dpushp, xu14, lantao\}@iu.edu}. \newline
    }
}

\conf{\hspace*{-9cm}\textcolor{gray}{This paper has been accepted for publication at the IEEE/RSJ International Conference on Intelligent Robots and Systems \textcolor{blue}{\href{https://ieee-iros.org/}{(IROS)}}, Detroit, Michigan, USA, 2023. }}

\begin{document}
\maketitle
\thispagestyle{empty}
\pagestyle{empty}

\begin{abstract}
Biological agents, such as humans and animals, are capable of making decisions out of a very large number of choices in a limited time. They can do so because they use their prior knowledge to find a solution that is not necessarily optimal but good enough for the given task. In this work, we study the motion coordination of multiple drones under the above-mentioned paradigm, Bounded Rationality (BR), to achieve cooperative motion planning tasks. Specifically, we design a prior policy that provides useful goal-directed navigation heuristics in familiar environments and is adaptive in unfamiliar ones via Reinforcement Learning augmented with an environment-dependent exploration noise.
Integrating this prior policy in the game-theoretic bounded rationality framework allows agents to quickly make decisions in a group considering other agents' computational constraints. 
Our investigation assures that agents with a well-informed prior policy increase the efficiency of the collective decision-making capability of the group. We have conducted rigorous experiments in simulation and in the real world to demonstrate that the ability of informed agents to navigate to the goal safely can guide the group to coordinate efficiently under the BR framework. 
\end{abstract}

\section{Introduction}
\label{sec:intro}

Efficient decision-making algorithms that can operate under computational constraints are essential in developing robotic systems. 
These systems need to select reasonable decisions from an extensive range of options within a limited time frame while constrained by finite computational resources. 
In multi-agent scenarios, such as generating motion trajectories that coordinate multiple agents, the challenge is further amplified, as the trajectories of different agents are interdependent, and the decisions made by one agent can impact the decisions of others. 
Moreover, when agents have different goals, balancing individual and collective decision-making objectives while considering each agent's computational limitations can make the problem even more complex. 

While game theory is a powerful framework for describing multi-agent interaction and computing interdependent motion trajectories that reach Nash Equilibrium among agents~\cite{osborne2004introduction, wang2019game}, it assumes that each agent is rational and has unlimited computational limits.
This assumption can be limiting in practical applications, as it may not accurately reflect the finite computational constraint of real-world agents. 
This limitation can be alleviated by combining it with information-theoretic bounded rationality (BR)~\cite{ortega2015information}.
It explicitly models each agent's computational limits by constraining the amount of information an agent is allowed to process when transitioning from a default policy, which  describes the nominal behavior of an agent before it initiates the planning process, to an optimal one~\cite{kappen2012optimal, MABRA_DARS2022, Ortega_IUBR}.
This transformation is typically computed via importance sampling~\cite{bishop2006pattern}, which samples a set of trajectories from the default policy to estimate the expected bounded-rational trajectories for each agent.


The performance of the resulting trajectory is highly dependent on the quality of the samples from the default policy. 
Previous works in the game-theoretic BR framework have typically used samples from uninformative distributions, such as a uniform or univariate Gaussian distribution~\cite{mohamed2022autonomous, williams2017autonomous, MABRA_DARS2022}, to estimate the expected bounded-rational trajectories for individual agents across all potential trajectories.
However, this ``brute-force" approach may be inefficient compared with the scenario when agents are equipped with certain prior knowledge about how to navigate in the environment. 
We are inspired by the biological agents who can leverage their prior knowledge to select reasonably good decisions, although suboptimal with limited processing effort. 
To achieve a similar mechanism, we propose two criteria that characterize a reasonable default policy -- informativeness and adaptiveness.
Informativeness emphasizes that the default policy should provide prior knowledge to guide the sampling process toward regions of higher utility, while adaptiveness ensures that the default policy explores additional regions when it underperforms.
To accomplish this, we leverage Goal-Conditioned RL~\cite{nasiriany2019planning} to train a goal-directed policy and use it as the default policy in the bounded-rational game-theoretic framework. 
Additionally, an environment-dependent exploration noise is added to the policy to allow exploration in underperforming environments. 
This noise provides additional variance to the agent's actions and allows them to deviate from their default policy to search for better trajectories when necessary. 
We integrate this default policy into the best response algorithm under the bounded rational framework.
By providing a reasonable guide for the sampling process, our algorithm can reduce the trajectories needed to optimize and predict agents' trajectories, resulting in a more efficient computational process overall. 

Our experimental results from simulations and physical experiments provide strong evidence that incorporating an informed prior policy within the BR framework reduces the number of candidate choices that need to be evaluated while promoting intelligent group behavior.
Our study reveals noteworthy findings concerning the improved performance of uninformed agents when working collaboratively with informed agents, which strikingly mirrors real-world organizational behavior. 
As shown in the experiments, the presence of informed agents in the group significantly enhances the decision-making efficiency of uninformed ones, as the informed default behaviors allow them to select decisions that not only optimize their objectives but also guide the less informed towards areas with higher rewards.
Thus, overall group performance is significantly enhanced.
These findings underscore the relevance and applicability of our research to real-world scenarios, where informed and uninformed agents synergistically contribute to enhanced group dynamics and overall performance.

\section{Related Work}
\label{sec:related_work}
Our research is relevant to the field of planning in multi-agent systems and game-theoretic frameworks. Here, we present a brief summary of these areas.


The game-theoretic framework has been widely used to model the interactions among multiple agents with different goals~\cite{owen2013game}. One approach to obtain a numerical solution for computing Nash Equilibrium is the iterative best response algorithm, which has been utilized in various multi-robot applications such as racing~\cite{wang2019game, spica2020real} and trajectory prediction and motion planning in self-driving vehicles~\cite{fisac2019hierarchical, schwarting2018planning, chen2017socially, lutjens2019safe}. 
In the context of uncertain environments, ~\cite{ogunmolu2018minimax} proposed a robust version of the algorithm by solving the minimax objective. It is worth noting, however, that the game-theoretic framework is based on the assumption of rational agents with unlimited computational power. 
This assumption may not hold in real-world scenarios, and thus other modeling approaches may be required.


Recently, there has been a growing interest in bounded rationality models \cite{Simon1990} that aim to formalize decision-making with limited resources, such as time, energy, memory, and computational effort. 
While several works have addressed the idea of terminating computation when time runs out, as documented in~\cite{ingrand2017deliberation}, only a handful have made an explicit attempt to model this bounded rational assumption. A recent study \cite{pacelli2021robust} examines the performance of single-agent robust control using information-theoretic bounded rationality, which has been previously explored in the works of Ortega and Braun~\cite{ortega2015information} and Genewein et al.~\cite{genewein2015bounded}. Additionally, a related but independently developed literature known as KL-Control has been studied in works such as Botvinick et al.\cite{botvinick2012planning} and Kappen et al.~\cite{kappen2012optimal}. The KL-divergence between a prior and a posterior policy is added as an additional constraint during the computation of the optimal policy. However, most of the work has been limited to single-agent problems. \cite{MABRA_DARS2022} extends information-theoretic bounded rationality to multi-agent problems by incorporating an iterated best response algorithm to find Nash equilibrium in a Markov game. However, it does not address the issue of an uninformative default policy. Although the proposed solution is motivated by biological agents naturally making complex decisions by selecting the satisfying option from a set of good options, this work does not address how to identify the good options.
The identification of a set of good options is highly dependent on prior knowledge of the task. Therefore, in this work, we address this issue by incorporating an informed prior policy into a multi-agent bounded rational decision-making framework.


\section{Background}

\subsection{Markov Games for Multi-Agent Trajectory Generation}
We consider the problem of generating motion trajectories in an environment consisting of multiple agents, where each agent's trajectory depends on and affects others.
This problem can be formulated using the framework of Markov Games~\citep{littman1994markov}, which extends Markov Decision Processes~\citep{bellman1966dynamic} to consider interactions among multiple agents.
A Markov Game is defined by a tuple $(\mathcal{N}, \{\mathcal{S}^i\}_{i \in \mathcal{N}}, \{\mathcal{A}^i\}_{i \in \mathcal{N}}, \{f^i\}_{i \in \mathcal{N}}, \{R^i\}_{i \in \mathcal{N}})$, where $\mathcal{N} = \{1, ..., N\}$ denotes the set of $N$ agents' indices; $s^i \in \mathcal{S}^i$ and $a^i \in \mathcal{A}^i$ denote the state and action spaces of agent $i$, respectively.
We use bold letters to denote agents' joint variables, e.g., joint states and actions are denoted as $\bm{s} = \{s^1, ..., s^N\}$ and $\bm{a} = \{a^1, ..., a^N\}$.
In this paper, we assume that each agent's transition function is deterministic and independent of others' states and actions, i.e., $f^i: \mathcal{S}^i \times \mathcal{A}^i \rightarrow \mathcal{S}^i$.
The reward functions $R^i: \mathcal{S} \times \mathcal{A} \rightarrow \mathbb{R}$ specifies the immediate reward of agent $i$ after all the agents take actions $\bm{a}$ at state $\bm{s}$.

At every timestep $t$, each agent $i\in \mathcal{N}$ simultaneously samples a sequence of actions until the planning horizon $H$ from their policies $A_t^i \sim \pi^{i}_{t}(A_t^i | \bm{s}_t, \bm{\pi}^{-i}_t)$, where $A_t^i = \{a_{t+1}^i, a_{t+2}^i, ..., a_{t+H-1}^i\}$, $\bm{s}_t$ is the agents' joint state at time $t$, and $\bm{\pi}^{-i}_t$ denotes the joint policy excludes agent $i$.
The conditioning of other agents' policies is necessary because each agent's reward and transition functions depend on the joint state and actions. 
The sequence of actions is then used to simulate the agents' states forward, generating state trajectories $S^i_{t} = \{s^i_{t}, s^i_{t+1}..., s^i_{t+H}\}$, where the next state is sampled based on the transition function $s^i_{t+k+1} = f^i({s}_{t+k}, {a}_{t+k})$.
This policy form is well-suited for most robotics problems because a low-level controller can be used to track the generated motion trajectories induced by the action sequence at every timestep~\cite{Mellinger2011}.
The goal for each agent is to find a policy $\pi_t^i$ that maximizes its expected long-term reward (or the utility function) given the joint state
\begin{equation}\label{eq:utility}
\pi_t^{i,*} = \textstyle{\argmax_{\pi_t^i}}U^{i}(\bm{s}_t, \bm{\pi}_t), 
\end{equation}
where $U^{i}(\bm{s}_t, \bm{\pi}_t)=\mathbb{E}\Big[\sum_{k=t}^{H+t} R^i(\bm{s}_k, \bm{a}_{k})\Big]$ is agent $i$'s utility function at $t$, $\bm{\pi}_t = \{\pi_{t}^{i}\}_{i \in \mathcal{N}}$ is the joint policy of all the agents, and $\bm{a}_{k}$ is the joint action at time $k$ extracted from the action sequence sampled from their corresponding policies.

\subsection{Solution Concept in Markov Games}
Since the utility function of each agent in a multi-agent setting depends not only on its own decisions but also on the actions taken by other agents, it is inadequate to optimize each agent's utility in isolation. 
Consequently, when optimizing the policies defined in Eq.~\eqref{eq:utility}, each agent must also consider the potential actions of other agents and how they may react to each other's decisions.
A commonly used solution concept to capture this interdependency among multiple interacting agents is Nash Equilibrium (NE)~\cite{osborne2004introduction} defined as follows:
\begin{align}\label{eq:best-response}
    U^i(\bm{s}_t, \bm{\pi}_t^{-i,*}, \pi_t^{i,*}) 
    \geq 
    &U^i(\bm{s}_t, \bm{\pi}_t^{-i,*}, \pi_t^{i}), \\ \nonumber
    &\forall i \in \{1, ..., N\},~
    \forall \pi_t^{i} \in \Pi^i, 
\end{align}
where $\bm{\pi}_t^{-i}$ denotes the joint policy without agent $i$ and $\Pi^i$ is the space of agent $i$'s policy. 
Intuitively, Eq.~\eqref{eq:best-response} suggests that agents cannot increase their utility through unilateral policy changes in equilibrium. 
To compute NE, a numerical technique called Iterative Best Response (IBR) can be applied~\cite{reeves2012computing}. 
This process begins with an initial estimate of the policies for all agents, which are then updated iteratively for each agent by determining the best response to the current trajectories of all other agents. 
It continues until convergence, i.e., no further changes occur in the joint policy. 

\section{Methodology}

\subsection{Information-Theoretic Bounded Rationality in Markov Games}
\label{sec:multi-agent-br}
To exactly solve the decision problem in Eq.~\eqref{eq:utility}, even restricted to a single agent, is intractable, as it requires enumerating all possible action sequences (usually infinite) and selecting the optimal one. 
This intractability is exacerbated when multiple interacting agents are involved.
In such scenarios, the agent must not only optimize its action sequence but also anticipate other agents' unknown behaviors (action outcomes). 
This requirement translates into the need to solve a single-agent decision problem for each agent multiple times, as reflected by the best response numerical solution stated in Eq~\eqref{eq:best-response}.
Consequently, the assumption of perfect rationality, commonly adopted in the standard game-theoretic framework, typically cannot be realized in the real world, as agents are constrained by their computational resources, forcing them only to evaluate a limited and finite number of trajectories to make a decision.

Instead, information-theoretic bounded rationality offers a more realistic modeling framework by explicitly considering agents' computational limits to trade-off between the available computational resources and the number of trajectories for evaluation~\cite{ortega2013thermodynamics, genewein2015bounded, MABRA_DARS2022}. 
This is achieved by defining a constraint considering the amount of information (computation) the agent can afford to transfer from a {\em default} (or prior) policy $q^i$ to the optimal one. 
Intuitively, this default policy describes the nominal behavior of the agent before initiating the planning process, and 
the bounded-rational agent seeks to identify an optimized posterior policy that falls within the neighborhood of the default policy.  
One might view the amount of available computational resources as an affordable amount of information. 
It is also referred to as uncontrolled ``free" dynamics, as previously discussed in~\cite{kappen2012optimal}. 

Formally, following the work~\cite{kappen2012optimal, MABRA_DARS2022}, we use KL-divergence to characterize this information-processing constraint
\begin{equation}\label{eq:constrained-value-fn}
\begin{split}
    \pi^{i, *}_t = \argmax_{\pi^{i}_t} U^i(\bm{s}_t, \bm{\pi}_t) 
    \textrm{, s. t. } D_{KL}(\pi^{i}_t || q^{i}_t) \leq K_i,
\end{split}
\end{equation}
where $D_{KL}$ is the KL divergence between the two stochastic policies, and $K_i$ is a constant denoting the amount of information (measured in bits) agent $i$ can deviate from the default policy. 
To optimize the policy under the constraint in Eq.~\eqref{eq:constrained-value-fn}, we first rewrite the constrained optimization problem as an unconstrained one
$\pi^{i, *}_t = \argmax_{\pi^i_t} U^i(\mathbf{s}_t, \bm{\pi}_t) - \frac{1}{\beta^i} KL(\pi^{i}_t || q^{i}_t)$, 
where $\beta_i > 0$ indicates the {\em rationality level}.
As shown in~\cite{MABRA_DARS2022}, the unconstrained problem is equivalent to 
\begin{equation}\label{eq:proof-distribution}
\pi^{i, *}_t = \argmax_{\pi^{i}_t} \Big\{-\frac{1}{\beta^i}D_{KL}(\pi^i_t || \psi^i_t) \Big\},
\end{equation}
where $\psi^i_t(A^i_t | \bm{s}_t) \propto q^{i}_t(A^i_t)e^{\beta^i U^i(\bm{s}_t, \bm{\pi}_t)}$.
The maximization of $-\frac{1}{\beta^i}D_{KL}(\pi^i_t||\psi^i_t)$ occurs only when the two distributions are equal.
As a result, while keeping other agents' policies fixed, the bounded-optimal policy is
\begin{equation}\label{eq:br-optimal}
    \pi^{i, *}_t(A^i_t | \bm{s}_t, \bm{\pi}^{-i}_t) = \frac{1}{Z^i_t} 
    q^i(A^i_t)e^{\beta^i U^i(\bm{s}_t, \bm{\pi}_t)},
\end{equation}
where $Z^i_t = \int q^i(A^i_t)e^{\beta^i\cdot U^i(\bm{s}_t, \bm{\pi}_t)} dA^i_t$ denotes the normalization constant.
When $\beta^i \rightarrow 0$, the bounded-rational policy is equivalent to the prior, reflecting the agent's inability to leverage computational resources. 
Conversely, when $\beta_i \rightarrow \infty$, the agent's decision-making becomes entirely rational and ignores the influence of the default policy. 
Compared to the optimal policy expressed in Eq.~\eqref{eq:utility}, this bounded-optimal policy enables a trade-off between performance and computational complexity via changing the rationality level $\beta^i$. 

\subsection{Efficient Bounded-Rational Trajectories Computation via Informative Default Policy}

\subsubsection{Bounded-Rational Policy Computation}
Selecting an action sequence using the bounded-optimal policy (Eq.\eqref{eq:br-optimal}) is equivalent to querying a sample from this distribution.
In this work, we build upon the importance sampling strategy proposed in~\cite{MABRA_DARS2022}, which estimates the expected action sequence of the distribution Eq.~\eqref{eq:br-optimal}.
The core idea is that since the default policy is generally easy to sample, each agent can compute the expected bounded-optimal action sequence by a weighted average over samples generated from their default policies, where the weights depend on the corresponding action sequence's utility. 
Specifically, at the best response iteration $l$, suppose that agent $i$'s predictions over other agents' action sequences as $\Bar{\bm{A}}_t^{-i,l}$, where the upper bar denotes the expected value. 
To update agent $i$'s best response to these predictions, we first sample $d \in \mathcal{D}, \mathcal{D} = \{1, ..., D \}$ action sequences from the default policy $\{\Delta A^{i, l}_{t, d} | 
 \Delta A^{i, l}_{t, d} \sim q^i_t\}_{d \in \mathcal{D}}$
They are then used to generate perturbed action sequences around the current mean estimates of agent $i$'s best response $A_{t,d}^{i,l} = \Bar{A}^{i,l}_{t,d} + \Delta A^{i,l}_{t,d}$. 
Combined with the current estimates of other agents' mean action sequences $\Bar{\bm{A}}^{-i,l}_{t}$,  are used to compute the mean for the next best response iteration $l+1$ using the importance sampling strategy 
\begin{equation}\label{eq:br-importance-sampling}
\begin{split}
    \Bar{A}^{i,l+1}_{t}
           \approx \frac{\sum_{d=1}^{D} w(A^{i,l}_{t,d}, \Bar{\bm{A}}^{-i,l}_{t})A^{i,l}_{t,d}}{\sum_{d=1}^{D} w(A^{i,l}_{t,d}, \Bar{\bm{A}}^{-i,l}_{t})},
\end{split}
\end{equation}
where $\Bar{A}^{i,l+1}_{t} = \mathbb{E}[A^{i, l+1}_{t} | \bm{s}_t, \bm{\pi}^{-i}_t]$ denotes the expected bounded-action sequence for agent $i$ at the $l+1$ best response iteration, and the weights are based on the trajectory utility given the perturbed action sequence and other agents' mean estimates $w(A^{i,l}_{t,k}\Bar{\bm{A}}^{-i,l}_{t}) = \exp\{\beta^i \hat{U}^{i,l}_{t,k}\}$.
The trajectory utility $\hat{U}^{i,l}_{t,k}$ is computed analogously to Eq.~\eqref{eq:utility}, which roll out the agents' dynamics based on the given action sequences $A^{i,l}_{t,d}$ and $\Bar{\bm{A}}^{-i,l}_{t}$.
The above update rule is integrated into the best response algorithm. 
It computes the bounded-rational Nash Equilibrium until a pre-defined number of iterations is reached or the policies converge.

\subsubsection{Informative Default Policy Design}
The perturbation strategy over the current mean estimate is critical for the final performance. 
Typically, it is achieved by adding noisy actions from an uninformative default policy $q_t^i$, e.g., a uniform distribution over all possible action sequences, to the mean estimation for each agent, as suggested in previous works~\cite{kappen2012optimal}. 
While this approach ensures perturbing the action sequences unbiasedly, as all possible action sequences are covered with equal probability, it may be less efficient when heuristics are available to guide the process.
Under this circumstance, this can lead to the need to sample more action sequences than necessary. 
To address this issue, we propose two criteria for designing a more informative default policy to increase computational efficiency.

\begin{algorithm}[t] 
\caption{{Bounded-Optimal Trajectories Computation via Informative Default Policies}}
\label{alg:best-response}
\begin{algorithmic}[1]
    \INPUT{
    Each agent default policies $\{q^i_{t}\}_{i \in \mathcal{N}}$;
    Transition functions $\{f^i\}_{i \in \mathcal{N}}$;
    Reward functions $\{R^i\}_{i \in \mathcal{N}}$;
    Rationality parameter $\{\beta^i\}_{i \in \mathcal{N}}$;
    Agents' current state $\bm{s}_t$;
    Each agent's goal $\bm{g}$;
    Number of action sequence samples $D$;
    Number of agents $N$;
    Number of best response iteration $N_{IBR}$;
    Number of action updates $N_{UP}$;
    }
    \OUTPUT{The expected bounded-optimal action sequences of each agent $\bm{\Bar{A}}^{*}_{t}$.}
    
    \State Initialize the mean of each agent to $0$, $\bm{\Bar{A}}_{t}^{0} \leftarrow 0$ .
    \For{$l \in \{1, ..., N_{IBR}\}$}
        \For{$i \in \mathcal{N}$}
            \For{$n \in \{1, ..., N_{UP}\}$}
                \State Assign initial states $s_{k,d}^{i,l}\leftarrow s^i_k, d \in \mathcal{D}$
                \State \parbox[t]{\dimexpr\linewidth-\algorithmicindent}
                {Rollout other agents' states based on $\bm{A}^{-i,l}_{t}$ \\ and $\bm{s}_t^{-i}$}
                \State \parbox[t]{\dimexpr\linewidth-\algorithmicindent}
                {// Adding guided exploration action to the current \\ action sequence}
                \For{$d \in \mathcal{D}$}
                    \State Initialize action sequence utility $U^{i,l}_{t, d} \leftarrow 0$
                    \For{$k \in \{t, ..., t+H-1\}$}
                        \State $\Delta a_{k, d}^{i,l} = q^i(\Delta a_{k, d}^{i,l} | s_{k, d}^{i,l}, g^i)$
                        \State 
                        {$a_{k,l}^{i, d} = \Delta a_{k,l}^{i, d} + \Bar{a}_{k}^{i,l} + \epsilon^i$}
                        \State $U^{i,l}_{t,d} \leftarrow U^{i,l}_{t,d} + R^i(\bm{s}^{l}_{k,d}, \bm{a}^l_{k,d})$
                        \State $s_{k+1,d}^{i,l} = f^i(s^{i,l}_{k,d}, a_{k,d}^{i,l})$
                    \EndFor
                \EndFor
                \State // Update $i^{th}$ agent mean action sequence
                \State $\Bar{A}^{i, l}_{t} = \frac{\sum_{d=1}^{D} w(A^{i,l-1}_{t,d}, \Bar{\bm{A}}^{-i,l-1}_{t})A^{i,l-1}_{t,d}}{\sum_{d=1}^{D} w(A^{i,l-1}_{t,d}, \Bar{\bm{A}}^{-i,l-1}_{t})}$ 
            \EndFor
        \EndFor
    \EndFor
    \State \textbf{return} $\Bar{A}^{i,N_{IBR}}_t$ for every agent $i \in \mathcal{N}$
  \end{algorithmic}
\end{algorithm} 
The first criterion is to ensure that the sampled action sequences concentrate on high-utility regions, meaning they should yield higher utility than an uninformative one.
In other words, the distance, e.g., KL divergence, between the informed policy and the optimal one is smaller so that a smaller sample size can be used to converge.
While constructing a generalized informed policy that works well in all possible environments is challenging, generating a policy in a specific environment can be more approachable. 
To achieve this, we leverage a task-conditioned RL policy trained using Proximal Policy Gradient (PPO)~\cite{schulman2017proximal}. 
The trained policy takes the navigation goal $g^i$ for agent $i$ and the agent's state $s^i_t$ as input and samples the best action $a_{t}^{i} \sim q^i_{t}(A_{t}^{i} | s^i_t, g^i)$ in a single-agent environment. 
In contrast to the uninformative policy, which does not consider any environments or tasks, this default policy is closed-loop.
Thus, to integrate the policy into the best-response bounded-rational framework, at each best-response iteration $l$, we compute the weight in Eq.~\eqref{eq:br-importance-sampling} by rolling out the trajectory through perturbing the mean action based on the closed-loop policy.

Specifically, the agent $i$'s perturbed action is given by $a_{k,l}^{i, d} = \Delta a_{k,l}^{i, d} + \Bar{a}_{k}^{i,l}$, where $\Bar{a}_{k}^{i,l}$ denotes $i^{th}$ agent's mean of the bounded-optimal action distribution at the $l^{th}$ best response iteration, and $\Delta a_{k, d}^{i,l} \sim q^i(\Delta a_{k, d}^{i,l} | s_{k, d}^{i,l}, g^i)$ is the noisy action sampled from the informed default policy. 
At the same time, the reward function is computed to accumulate the utility, which is used to calculate the weight. 
This computation is performed along the action sequence until the end of the horizon by rolling out next states $s_{k+1,d}^{i,l} = f^i(s^{i,l}_{k,d}, a_{k,d}^{i,l}), k \in \{t, ..., t+H-1\}$. 
The informative policy guides this perturbation strategy by directing the agent's actions toward the goal, providing more efficient sampling.

The default policy for goal-point navigation can be viewed as a useful heuristic, but its performance is limited to the trained environment. 
In new environments, using samples from the default policy may perform worse than those from the uninformed one. 
To address this issue, we propose a second criterion that introduces diversity into the generated action sequences, covering wider regions in environments where the default policy may not perform reasonably.
To achieve this, we augment the default policy with an additional noise term, denoted by $\Delta a^{i,l}_{k,d} + \epsilon^i$, where $a$ is the sampled action, and $\epsilon^i$ is the added noise for agent $i$. The spread of the noise, such as the variance in the Gaussian distribution, controls the degree of adaptation of the default policy, with larger ranges facilitating the agent's performance in unfamiliar environments but requiring more samples to converge.
Incorporating this exploration noise is necessary for deploying in unfamiliar environments, and it enhances the flexibility and effectiveness of the importance sampling strategy in adapting to different scenarios. 
This is examined by varying the range of the noise parameter on the performance of the bounded-rational policy, shedding light on the trade-off between adaptability and noise variance.

The final algorithm for each agent's computation process is illustrated in Alg.~\ref{alg:best-response}. 
To compute the bounded-optimal action sequence of agent $i$ at the $l^{th}$ best response iteration, we first rollout the trajectories of other agents based on their current mean estimates $\bm{A}^{-i, l}_t$. 
Then, while keeping other agents' trajectories fixed from line 8 to line 15, we compute utilities of $D$ sampled trajectories of agent $i$ by incorporating the guided action noise into the mean. 
These utilities are used in Eq.~\eqref{eq:br-importance-sampling} to update the expected action sequence of agent $i \in \mathcal{N}$ at the best response iteration $l$.
This best response computation is carried out until a pre-defined number of iterations is reached. 

\vspace{5pt}
\section{Simulation Experiments}
\vspace{-0.3cm}
\begin{figure}[h]
    \centering
    \centering
  \subfigure[Training]
  	{\includegraphics[height=0.6in, width=0.75in]
    {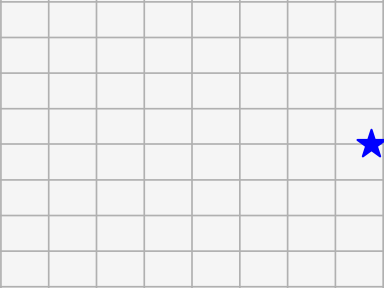}
         \label{fig:env0}}
  \subfigure[Test$(Env1)$]
  	{\includegraphics[height=0.6in, width=0.75in]
    {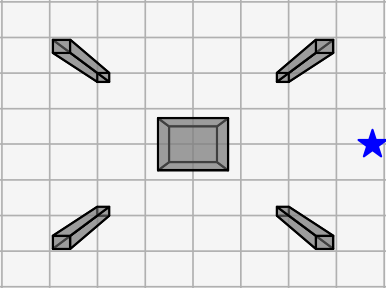}
         \label{fig:env1}}
  \subfigure[Test$(Env2)$]  	
    {\includegraphics[height=0.6in, width=0.75in]
    {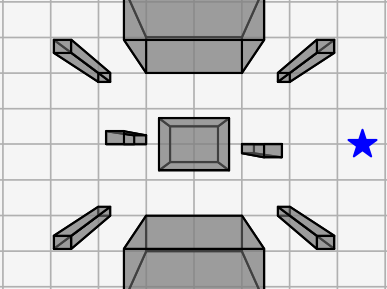}
         \label{fig:env2}}
  \subfigure[Test$(Env3)$]
  	{\includegraphics[height=0.6in, width=0.75in]
    {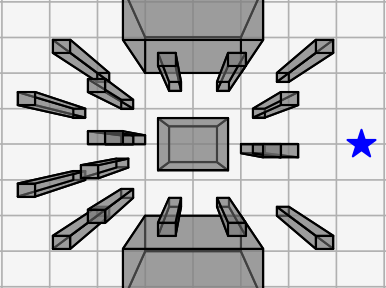}
         \label{fig:env3}}
    \caption{Shows the training and testing environments. Star represents the pre-defined goal position during training and testing single agent behavior.}    
    \label{fig:environments}
\end{figure}
\label{sec:sim-experiment}
\begin{figure*}[t!] 
  \centering
  \subfigure[Performance vs Sampled Trajectories]
  	{\includegraphics[height=1.4in, width=2.2in]
    {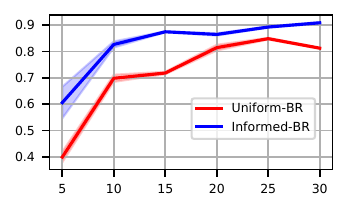}
         \label{fig:A1}}
  \subfigure[Performance vs Test Environments]
  	{\includegraphics[height=1.4in, width=2.2in]
    {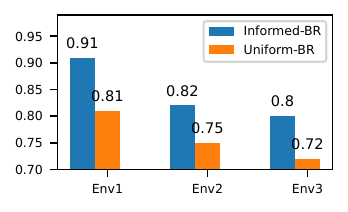}
         \label{fig:A2}}
  \subfigure[Performance vs Variance]  	
    {\includegraphics[height=1.4in, width=2.2in]{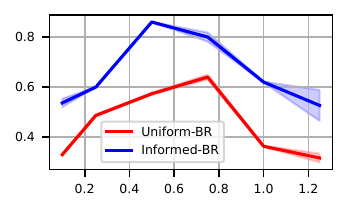}
         \label{fig:A3}}
  \caption{The performance comparison across different training environments. Y-axis indicates the normalized scores while the X-axis indicates the number of sampled trajectories in (a) and the environment density level in (b), ranging from sparse (Env1) to dense (Env3) as depicted in Fig.~\ref{fig:environments}. (c) shows the performance by changing variance of the prior distribution.
  \vspace{-10pt}
  }
\label{fig:A}  
\end{figure*}

\begin{figure*}[t!] 
  \centering
  \subfigure[Informed-BR $(t_1)$]
  	{\includegraphics[height=0.9in, width=1.3in]
    {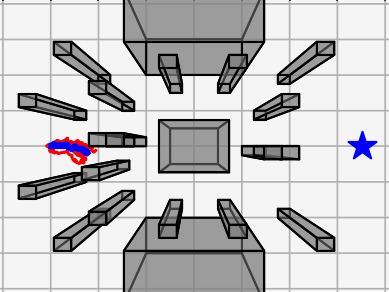}
         \label{fig:Inf_t1}}
  \subfigure[Informed-BR $(t_2)$]
  	{\includegraphics[height=0.9in, width=1.3in]
    {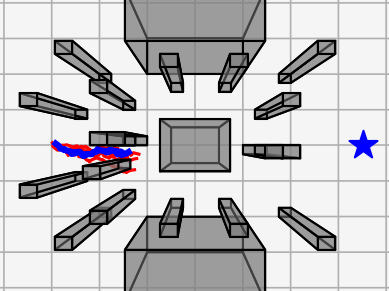}
         \label{fig:Inf_t2}}
  \subfigure[Informed-BR $(t_3)$]  	
    {\includegraphics[height=0.9in, width=1.3in]{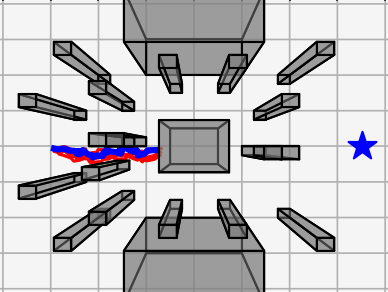}
         \label{fig:Inf_t3}}
  \subfigure[Informed-BR $(t_4)$]
  	{\includegraphics[height=0.9in, width=1.3in]
    {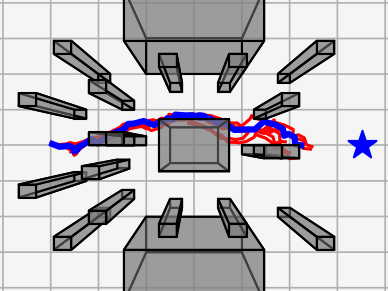}
         \label{fig:Inf_t3}}
  \subfigure[Informed-BR $(t_5)$]
  	{\includegraphics[height=0.9in, width=1.3in]
    {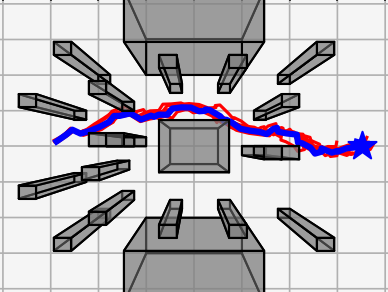}
         \label{fig:Inf_t5}}
  \subfigure[Uniform-BR $(t_1)$]
  	{\includegraphics[height=0.9in, width=1.3in]
    {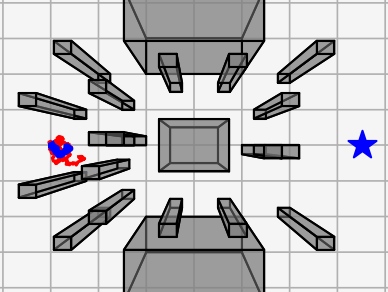}
         \label{fig:U_t1}}
  \subfigure[Uniform-BR $(t_2)$]
  	{\includegraphics[height=0.9in, width=1.3in]
    {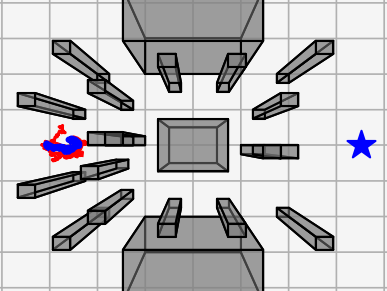}
         \label{fig:U_t2}}
  \subfigure[Uniform-BR $(t_3)$]  	
    {\includegraphics[height=0.9in, width=1.3in]{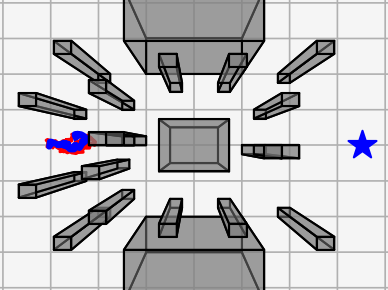}
         \label{fig:U_t3}}
  \subfigure[Uniform-BR $(t_4)$]
  	{\includegraphics[height=0.9in, width=1.3in]
    {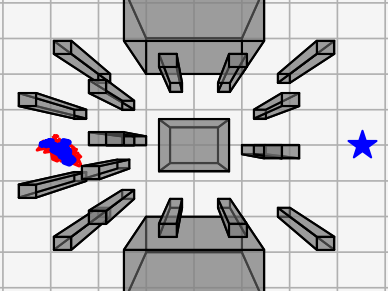}
         \label{fig:U_t3}}
  \subfigure[Uniform-BR $(t_5)$]
  	{\includegraphics[height=0.9in, width=1.3in]
    {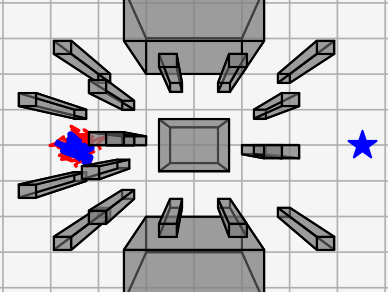}
         \label{fig:U_t5}}
  \caption{Shows the adaptation process from default to satisficing path in the proposed method and the baseline. (a) to (e) shows the proposed method's performance on the adaptability. (f) to (j) shows the baseline method's performance. The snapshot has been taken after the same number of iteration to shoe the comparison. Both the method uses $5$ sampled trajectories.  
  \vspace{-15pt}
  }
\label{fig:single_agent_path}  
\end{figure*}

We perform comprehensive simulation experiments to investigate the impact of incorporating prior knowledge about a task on the decision-making performance of agents operating within a Bounded Rationality framework. Our findings reveal that integrating an informed prior policy into the BR framework significantly reduces the number of candidate choices required to be evaluated for identifying a satisfying (i.e., satisfactory and sufficient) trajectory. In this section, we specifically demonstrate that utilizing an informed prior policy enables a single agent to (1) rapidly adapt to novel environments, and (2) reason about the behavior of other agents in a multi-agent system while requiring fewer trajectory samples than the uninformed one.

\subsection{Simulation Setup and Performance Metrics}
We designed two sets of experiments to validate the effectiveness of our proposed approach. The first experiment involves navigating a single robot in a 3D environment to a pre-defined goal point, while the second experiment involves navigating a group of aerial vehicles in a 3D space to a pre-defined goal while avoiding collisions with each other as well as obstacles in the environment. 

\textbf{Performance Metric: }We use the normalized score to compare performance across different environment setups. 
The score is calculated as $\frac{J(\tau^{\pi}) - J(\tau^{rand})}{J(\tau^{*}) - J(\tau^{rand})}$, where $\tau^{rand}$, $\tau^{\pi}$, and $\tau^{*}$ are the trajectories of a uniform policy, computed policy, and the expert policy computed by using a large number of trajectories, and $J(\cdot)$ is the utility of the trajectory. 
The larger the score, the better the performance, and a score of $0$ indicates that the performance of a computed policy is the same as the uniform policy.

\textbf{Environments: }We train a single agent in an obstacle-free environment and test the performance of its adaptability in $3$ different environments with increasing difficulties as shown in the Fig.~\ref{fig:environments}.

\begin{figure*}[t] 
  \centering
  \subfigure[Performance vs Sampled Trajectories]
  	{\includegraphics[height=1.4in, width=2.2in]
    {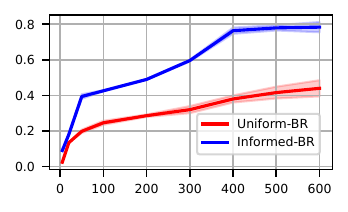}
         \label{fig:B1}}
  \subfigure[Performance vs Environments]
  	{\includegraphics[height=1.4in, width=2.2in]
    {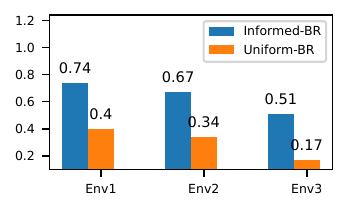}
         \label{fig:B2}}
  \subfigure[Agents Replacemet]  	
    {\includegraphics[height=1.4in, width=2.2in]{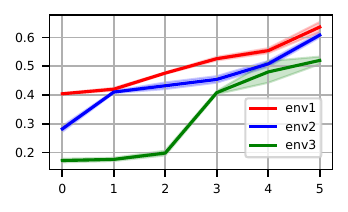}
         \label{fig:B3}}
  \subfigure[Group Performance (Env1)]
  	{\includegraphics[height=1.4in, width=2.2in]
    {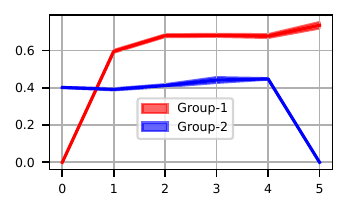}
         \label{fig:B3_a}}
  \subfigure[Group-1 performance]
  	{\includegraphics[height=1.4in, width=2.2in]
    {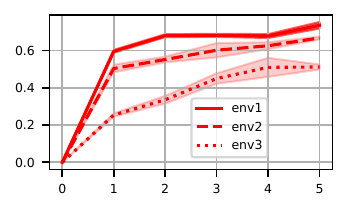}
         \label{fig:B3_b}}
  \subfigure[Group-2 performance]  	
    {\includegraphics[height=1.4in, width=2.2in]{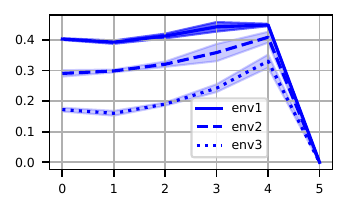}
         \label{fig:B3_c}}
  \caption{Illustration of the performance evaluation of five agents across different environments. (a) Shows the performance agents in Env1. (b) Shows the performance of agents on three different environments. Number of sampled trajectories used in all the environments and in all trials is $400$. In (c), (d), (e) and (f) the X-axis represents the number of Informed-BR agents in a 5-agent group, and the Y-axis represents the performance of the group. (c) Depicts the performance of replacing Uniform-BR agents with Informed-BR agents. (d) Shows the performance of two groups separately, where Group-1 is the group of all Informed-BR agents and Group-2 is the group of all Uniform-BR agents. (e) and (f) shows their performance across the testing environments.The performance of Group-1 and Group-2 is zero, when there are no agents assigned to those groups.
  }
\label{fig:multi_agent_exp}  
\end{figure*}

We intentionally select an obstacle-free training environment to illustrate the efficacy of the proposed method. However, it is imperative to note that a closer similarity between the training and testing environments is expected to yield superior performance results.

\textbf{Agent's Description and Fixed Parameters: } Homogeneous agents are used in both experiments with identical sizes and physical capabilities. The transition functions of the agents are set using a deterministic single integrator model with the minimum and maximum speeds as $a_{min} = 0 m/s$ and $a_{max} = 1m/s$. 
Rationality levels of all agents are assumed to be known in all simulations. The one-step reward function is designed to penalize collisions and large distances to the goal. Each simulation is run $10$ times for $T=80$ timesteps.

\textbf{Baseline: }The method employed as the baseline for comparison uses a uninformative uniform prior policy in the game theoretic bounded rationality framework, as described in the literature \cite{MABRA_DARS2022}.


\subsection{Adaptability of Single-Agent in Unknown Environment }
We first show that an {\em Informed} agent adapts faster in unknown environment. To compare their performance, we gradually increase the number of sampled trajectories from $5$ to $30$, while maintaining a constant level of rationality ($\beta=0.03$) and variance ($\sigma = 0.50$) as depicted in Fig.~\ref{fig:A1}. The informed policy method (Informed-BR) achieves $84\%$ performance by sampling $10$ trajectories. However, the uniform policy (Uniform-BR) agent needs more than $20$ trajectories on average to gain the same performance. The comparison of the performance of both methods across all environments, while keeping the number of sampled trajectories fixed at $30$ (where the Uniform-BR performs best), is presented in Fig.~\ref{fig:A2}. Our observations reveal that the Informed-BR agent outperforms the Uniform-BR in all environments, thus validating our proposal of employing Informed-BR. 
Fig.~\ref{fig:A3} shows the importance of variance in determining the informativeness of the default policy.  
  
Fig.~\ref{fig:single_agent_path} displays the generated trajectories for both the Informed-BR and Uniformed-BR methods at $5$ consecutive optimization iterations, i.e., $n \in \{1, ..., 5\}$ (see Algorithm~\ref{alg:best-response}). 
Despite only using $5$ trajectories, the Informed-BR method achieves a satisfactory performance compared to the Uniform-BR method.
The readers are encouraged to watch the experimental videos at \url{https://youtu.be/iD6xyyxQCGI}. 

\begin{figure*}[t!]
    \centering
    \centering
  \subfigure[Experiment-1 (set up)]
  	{\includegraphics[height=1.2in, width=1.7in]
    {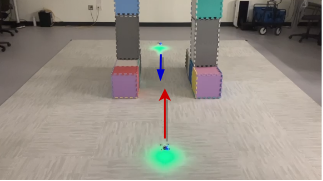}
         \label{fig:env0}}
  \subfigure[Experiment-1 (snapshot)]
  	{\includegraphics[height=1.2in, width=1.7in]
    {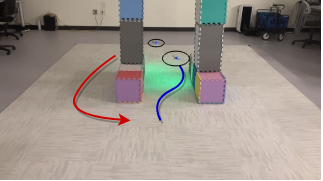}
         \label{fig:env1}}
  \subfigure[Experiment-2 (set up)]  	
    {\includegraphics[height=1.2in, width=1.7in]
    {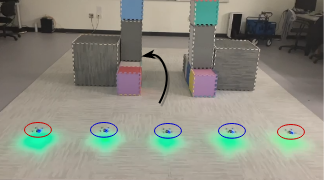}
         \label{fig:env2}}
  \subfigure[Experiment-2 (snapshot)]
  	{\includegraphics[height=1.2in, width=1.7in]
    {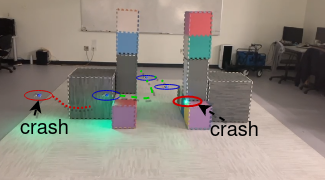}
         \label{fig:env3}}
    \caption{Shows the training and testing environments. Star represents the pre-defined goal position during training and testing single agent behavior.\vspace{-10pt}}    
    \label{fig:environments}
\end{figure*}

\subsection{Faster Adaptability in Multi-Agent Scenario}
After demonstrating the adaptability of a single {\em Informed-BR} agent in previously unseen environments, we now present the performance of multiple agents to further validate the superiority of our method over {\em Uniform-BR}. We added four additional agents in all three environments at various locations and assigned them the same goal as specified in Fig.~\ref{fig:environments}. The rationality levels $\beta_i$ of all agents were fixed at $0.1$ in all trials. Fig.~\ref{fig:B1} shows that when all agents are informed, they can achieve a normalized score of approximately $0.80$ with $400$ trajectories, while the Uniform-BR agents can only achieve half of that. This result indicates that the informed policy not only adapts to the static unseen environment faster than the Uniform-BR, but it can also adapt to avoid dynamic objects (other agents in this case) in multi-agent environments. 
Fig.~\ref{fig:B2} demonstrates the effect of increasing environmental complexity, measured by the space occupied by the obstacles, on the performance of Informed-BR and Uniform-BR methods. While both methods exhibit a decrease in performance as environmental complexity increases, the Informed-BR consistently outperforms the Uniform-BR in all environments. Specifically, in the most complex environment (Env3), the Informed-BR method achieves a normalized score of around 0.5 while the Uniform-BR method only achieves around 0.2, indicating the effectiveness of our method in adapting to complex environments.

In the final experiment, we investigated the impact of replacing Uniform-BR agents with Informed-BR agents on group performance. As illustrated in Fig.~\ref{fig:B3}, the heterogeneous group's performance gradually improves as more Uniform-BR agents are replaced with Informed-BR agents. Interestingly, replacing the uninformed agents with informed ones not only increases the performance of the informed group, but also induces a positive effect on the uniform group, as demonstrated by the increasing trend of the blue line in Fig.~\ref{fig:B3_a} and more clearly shown in Fig.~\ref{fig:B3_c}. Moreover, Fig.~\ref{fig:B3_b} shows that adding more informed members to the group enhances its overall performance. This effect is commonly observed in human organizations, where the addition of more intelligent agents can help overcome the limitations of bounded rationality. Our modelled agents exhibit the same behavior, as they are also boundedly rational. 
We posit 
that the presence of more informed agents in a group may positively affect uninformed agents' performance. We speculate that this could be due to the fact that informed agents reduce the search space for the other agents and their trajectories guide the uninformed agents in the right direction. However, further research is needed to confirm whether this phenomenon holds true in different scenarios and configurations of robot states.

\section{Physical Experiments}
In this section, we present the physical experiments conducted to validate the proposed method using Crazyflie 2.1 nano-drones under a motion capture system. 
Two tasks were considered for the hardware experiment. In the first task, two agents with informed and uniform prior policies were used to swap their positions while avoiding obstacles and each other, as shown in Fig.\ref{fig:env0}. Note that the narrow gap between the agents can only allow one agent at a time. The agent with the informed prior policy successfully took the shortest path through the narrow gap, while the agent with the uniform prior policy took a longer route to the destination, as shown in Fig.\ref{fig:env1}. This interesting behavior supports our simulation results and also shows the importance to having informed prior policy. \vspace{-1pt}

In the second task, a group of five nano-drones was selected and divided into two groups, namely Group-1 and Group-2. Group-1, indicated by the red circles in Fig.~\ref{fig:env2}, employed a uniform prior policy, while Group-2, shown by the blue circles, utilized an informed prior policy. The objective of both groups was to navigate through a narrow passage and reach the goal point. The experimental results demonstrate that the Uniform-BR agents failed to accomplish the task, whereas the Informed-BR agent succeeded. This is attributed to the insufficiency of the number of trajectories sampled from the default policy in the baseline, while the proposed method performs optimally.

\section{Conclusion}
This paper presents a novel approach to multi-agent motion coordination by leveraging Bounded Rationality and an informative prior policy. 
Our experiments demonstrate that using learned navigation strategies in familiar environments and adaptive exploration in unfamiliar ones can increase the efficiency of the collective decision-making capability of the group. Moreover, integrating the prior policy in the game-theoretic BR framework allows agents to make decisions while considering other agents' computational constraints quickly. 
Through various experiments in simulation and the real world, we validate our approach's effectiveness. To the best of our knowledge, this study is the first to investigate the use of a well-informed prior policy in coordinating multiple drones under the BR framework. Our findings suggest that the proposed approach has significant potential for designing efficient and adaptive swarm systems. Future research can explore the generalizability of our approach to other multi-agent systems and further improve its robustness to uncertainty and partial observability.

{\small
\bibliography{ref} 

\begin{thebibliography}{28}
\providecommand{\natexlab}[1]{#1}
\providecommand{\url}[1]{#1}
\csname url@samestyle\endcsname
\providecommand{\newblock}{\relax}
\providecommand{\bibinfo}[2]{#2}
\providecommand{\BIBentrySTDinterwordspacing}{\spaceskip=0pt\relax}
\providecommand{\BIBentryALTinterwordstretchfactor}{4}
\providecommand{\BIBentryALTinterwordspacing}{\spaceskip=\fontdimen2\font plus
\BIBentryALTinterwordstretchfactor\fontdimen3\font minus
  \fontdimen4\font\relax}
\providecommand{\BIBforeignlanguage}[2]{{%
\expandafter\ifx\csname l@#1\endcsname\relax
\typeout{** WARNING: IEEEtranN.bst: No hyphenation pattern has been}%
\typeout{** loaded for the language `#1'. Using the pattern for}%
\typeout{** the default language instead.}%
\else
\language=\csname l@#1\endcsname
\fi
#2}}
\providecommand{\BIBdecl}{\relax}
\BIBdecl

\bibitem[Osborne et~al.(2004)]{osborne2004introduction}
M.~J. Osborne \emph{et~al.}, \emph{An introduction to game theory}.\hskip 1em
  plus 0.5em minus 0.4em\relax Oxford university press New York, 2004, vol.~3,
  no.~3.

\bibitem[Wang et~al.(2019)Wang, Wang, Talbot, Gerdes, and
  Schwager]{wang2019game}
M.~Wang, Z.~Wang, J.~Talbot, J.~C. Gerdes, and M.~Schwager, ``Game theoretic
  planning for self-driving cars in competitive scenarios.'' in \emph{Robotics:
  Science and Systems}, 2019.

\bibitem[Ortega et~al.(2015)Ortega, Braun, Dyer, Kim, and
  Tishby]{ortega2015information}
P.~A. Ortega, D.~A. Braun, J.~Dyer, K.-E. Kim, and N.~Tishby,
  ``Information-theoretic bounded rationality,'' \emph{arXiv preprint
  arXiv:1512.06789}, 2015.

\bibitem[Kappen et~al.(2012)Kappen, G{\'o}mez, and Opper]{kappen2012optimal}
H.~J. Kappen, V.~G{\'o}mez, and M.~Opper, ``Optimal control as a graphical
  model inference problem,'' \emph{Machine learning}, vol.~87, no.~2, pp.
  159--182, 2012.

\bibitem[Xu et~al.(2022)Xu, Pushp, Yin, and Liu]{MABRA_DARS2022}
J.~Xu, D.~Pushp, K.~Yin, and L.~Liu, ``Decision-making among bounded rational
  agents,'' 2022.

\bibitem[Ortega and Braun(2011)]{Ortega_IUBR}
D.~A. Ortega and P.~A. Braun, ``Information, utility and bounded rationality,''
  in \emph{Artificial General Intelligence}, J.~Schmidhuber, K.~R.
  Th{\'o}risson, and M.~Looks, Eds.\hskip 1em plus 0.5em minus 0.4em\relax
  Berlin, Heidelberg: Springer Berlin Heidelberg, 2011, pp. 269--274.

\bibitem[Bishop and Nasrabadi(2006)]{bishop2006pattern}
C.~M. Bishop and N.~M. Nasrabadi, \emph{Pattern recognition and machine
  learning}.\hskip 1em plus 0.5em minus 0.4em\relax Springer, 2006, vol.~4,
  no.~4.

\bibitem[Mohamed et~al.(2022)Mohamed, Yin, and Liu]{mohamed2022autonomous}
I.~S. Mohamed, K.~Yin, and L.~Liu, ``Autonomous navigation of agvs in unknown
  cluttered environments: log-mppi control strategy,'' \emph{IEEE Robotics and
  Automation Letters}, vol.~7, no.~4, pp. 10\,240--10\,247, 2022.

\bibitem[Williams et~al.(2017)Williams, Goldfain, Drews, Rehg, and
  Theodorou]{williams2017autonomous}
G.~Williams, B.~Goldfain, P.~Drews, J.~M. Rehg, and E.~A. Theodorou,
  ``Autonomous racing with autorally vehicles and differential games,''
  \emph{arXiv preprint arXiv:1707.04540}, 2017.

\bibitem[Nasiriany et~al.(2019)Nasiriany, Pong, Lin, and
  Levine]{nasiriany2019planning}
S.~Nasiriany, V.~Pong, S.~Lin, and S.~Levine, ``Planning with goal-conditioned
  policies,'' \emph{Advances in Neural Information Processing Systems},
  vol.~32, 2019.

\bibitem[Owen(2013)]{owen2013game}
G.~Owen, \emph{Game theory}.\hskip 1em plus 0.5em minus 0.4em\relax Emerald
  Group Publishing, 2013.

\bibitem[Spica et~al.(2020)Spica, Cristofalo, Wang, Montijano, and
  Schwager]{spica2020real}
R.~Spica, E.~Cristofalo, Z.~Wang, E.~Montijano, and M.~Schwager, ``A real-time
  game theoretic planner for autonomous two-player drone racing,'' \emph{IEEE
  Transactions on Robotics}, vol.~36, no.~5, pp. 1389--1403, 2020.

\bibitem[Fisac et~al.(2019)Fisac, Bronstein, Stefansson, Sadigh, Sastry, and
  Dragan]{fisac2019hierarchical}
J.~F. Fisac, E.~Bronstein, E.~Stefansson, D.~Sadigh, S.~S. Sastry, and A.~D.
  Dragan, ``Hierarchical game-theoretic planning for autonomous vehicles,'' in
  \emph{2019 International Conference on Robotics and Automation (ICRA)}.\hskip
  1em plus 0.5em minus 0.4em\relax IEEE, 2019, pp. 9590--9596.

\bibitem[Schwarting et~al.(2018)Schwarting, Alonso-Mora, and
  Rus]{schwarting2018planning}
W.~Schwarting, J.~Alonso-Mora, and D.~Rus, ``Planning and decision-making for
  autonomous vehicles,'' \emph{Annual Review of Control, Robotics, and
  Autonomous Systems}, vol.~1, pp. 187--210, 2018.

\bibitem[Chen et~al.(2017)Chen, Everett, Liu, and How]{chen2017socially}
Y.~F. Chen, M.~Everett, M.~Liu, and J.~P. How, ``Socially aware motion planning
  with deep reinforcement learning,'' in \emph{2017 IEEE/RSJ International
  Conference on Intelligent Robots and Systems (IROS)}.\hskip 1em plus 0.5em
  minus 0.4em\relax IEEE, 2017, pp. 1343--1350.

\bibitem[L{\"u}tjens et~al.(2019)L{\"u}tjens, Everett, and
  How]{lutjens2019safe}
B.~L{\"u}tjens, M.~Everett, and J.~P. How, ``Safe reinforcement learning with
  model uncertainty estimates,'' in \emph{2019 International Conference on
  Robotics and Automation (ICRA)}.\hskip 1em plus 0.5em minus 0.4em\relax IEEE,
  2019, pp. 8662--8668.

\bibitem[Ogunmolu et~al.(2018)Ogunmolu, Gans, and Summers]{ogunmolu2018minimax}
O.~Ogunmolu, N.~Gans, and T.~Summers, ``Minimax iterative dynamic game:
  Application to nonlinear robot control tasks,'' in \emph{2018 IEEE/RSJ
  International Conference on Intelligent Robots and Systems (IROS)}.\hskip 1em
  plus 0.5em minus 0.4em\relax IEEE, 2018, pp. 6919--6925.

\bibitem[Simon(1990)]{Simon1990}
\BIBentryALTinterwordspacing
H.~A. Simon, \emph{Bounded Rationality}.\hskip 1em plus 0.5em minus 0.4em\relax
  London: Palgrave Macmillan UK, 1990, pp. 15--18. [Online]. Available:
  \url{https://doi.org/10.1007/978-1-349-20568-4_5}
\BIBentrySTDinterwordspacing

\bibitem[Ingrand and Ghallab(2017)]{ingrand2017deliberation}
F.~Ingrand and M.~Ghallab, ``Deliberation for autonomous robots: A survey,''
  \emph{Artificial Intelligence}, vol. 247, pp. 10--44, 2017.

\bibitem[Pacelli and Majumdar(2021)]{pacelli2021robust}
V.~Pacelli and A.~Majumdar, ``Robust control under uncertainty via bounded
  rationality and differential privacy,'' \emph{arXiv preprint
  arXiv:2109.08262}, 2021.

\bibitem[Genewein et~al.(2015)Genewein, Leibfried, Grau-Moya, and
  Braun]{genewein2015bounded}
T.~Genewein, F.~Leibfried, J.~Grau-Moya, and D.~A. Braun, ``Bounded
  rationality, abstraction, and hierarchical decision-making: An
  information-theoretic optimality principle,'' \emph{Frontiers in Robotics and
  AI}, vol.~2, p.~27, 2015.

\bibitem[Botvinick and Toussaint(2012)]{botvinick2012planning}
M.~Botvinick and M.~Toussaint, ``Planning as inference,'' \emph{Trends in
  cognitive sciences}, vol.~16, no.~10, pp. 485--488, 2012.

\bibitem[Littman(1994)]{littman1994markov}
M.~L. Littman, ``Markov games as a framework for multi-agent reinforcement
  learning,'' in \emph{Machine learning proceedings 1994}.\hskip 1em plus 0.5em
  minus 0.4em\relax Elsevier, 1994, pp. 157--163.

\bibitem[Bellman(1966)]{bellman1966dynamic}
R.~Bellman, ``Dynamic programming,'' \emph{science}, vol. 153, no. 3731, pp.
  34--37, 1966.

\bibitem[Mellinger and Kumar(2011)]{Mellinger2011}
D.~Mellinger and V.~Kumar, ``Minimum snap trajectory generation and control for
  quadrotors,'' in \emph{2011 IEEE International Conference on Robotics and
  Automation}, 2011, pp. 2520--2525.

\bibitem[Reeves and Wellman(2012)]{reeves2012computing}
D.~Reeves and M.~P. Wellman, ``Computing best-response strategies in infinite
  games of incomplete information,'' \emph{arXiv preprint arXiv:1207.4171},
  2012.

\bibitem[Ortega and Braun(2013)]{ortega2013thermodynamics}
P.~A. Ortega and D.~A. Braun, ``Thermodynamics as a theory of decision-making
  with information-processing costs,'' \emph{Proceedings of the Royal Society
  A: Mathematical, Physical and Engineering Sciences}, vol. 469, no. 2153, p.
  20120683, 2013.

\bibitem[Schulman et~al.(2017)Schulman, Wolski, Dhariwal, Radford, and
  Klimov]{schulman2017proximal}
J.~Schulman, F.~Wolski, P.~Dhariwal, A.~Radford, and O.~Klimov, ``Proximal
  policy optimization algorithms,'' \emph{arXiv preprint arXiv:1707.06347},
  2017.

\end{thebibliography}
\bibliographystyle{IEEEtranN}
}
\end{document}